\documentclass[letterpaper, 10 pt, conference]{ieeeconf}

\IEEEoverridecommandlockouts                              

\usepackage{graphicx}
\usepackage{amssymb}
\usepackage{pifont}
\usepackage[ruled,noend]{algorithm2e}
\SetKwInput{KwIn}{Input}
\SetKwInput{KwOut}{Output}
\SetKw{KwRet}{return}            
\SetKw{KwBreak}{break}   
\DontPrintSemicolon
\SetAlgoNoLine
\SetAlgoInsideSkip{2pt}
\SetAlgoSkip{4pt}
\SetInd{0.6em}{0.6em}
\usepackage{amsmath} 
\usepackage{url} 
\usepackage{booktabs}
\usepackage{hyperref}
\usepackage{multirow}
\usepackage{graphicx}
\usepackage{xcolor}
\usepackage{soul}
\usepackage{comment}
\usepackage{microtype}
\usepackage{seqsplit}
\usepackage{xurl}
\usepackage[most]{tcolorbox}

\newcommand{\ttb}[1]{\texttt{\nolinkurl{#1}}}

\newcommand{\cm}{\emph{ContextMatters}}
\newcommand{\eg}{\emph{e.g.,}\xspace}
\newcommand{\ie}{\emph{i.e.,}\xspace}

\newcommand{\ssem}{\mathcal{S}_{\text{sem}}}

\newcommand{\boldred}[1]{\textcolor{red}{\textbf{#1}}}
\newcommand{\boldblack}[1]{\textcolor{black}{\textbf{#1}}}
\newcommand{\boldblue}[1]{\textcolor{blue}{\textbf{#1}}}

\definecolor{gammashiftcolor}{RGB}{251, 231, 206}
\definecolor{VALcolor}{RGB}{145, 117, 164}
\definecolor{groundcolor}{RGB}{114, 141, 188}
\definecolor{relaxcolor}{RGB}{174, 195, 233}
\definecolor{plancolor}{RGB}{172, 92, 83}

\usepackage{algorithmicx} 
\usepackage{algpseudocode} 

\usepackage[table]{xcolor} 

\title{\LARGE \bf Context Matters!\\Relaxing Goals with LLMs for Feasible 3D Scene Planning
}

\author{Emanuele Musumeci$^{1,*}$, Michele Brienza$^{1,*}$, Francesco Argenziano$^{1,*}$, Abdel Hakim Drid$^{2}$, Vincenzo Suriani$^{1}$, \\ Daniele Nardi$^{1}$, and Domenico D. Bloisi$^{3}$%
\thanks{$^{1}$Department of Computer, Automation and Management Engineering,
        Sapienza University of Rome, Rome, Italy
        {\tt\small lastname@diag.uniroma1.it} 
        $^{2}$Department of Electronics and Automation, Mohamed Khider University of Biskra, Biskra, Algeria
        {\tt\small abdelhakim.drid@univ-biskra.dz}
        $^{3}$International University of Rome UNINT, Rome, Italy
        {\tt\small domenico.bloisi@unint.eu}
        $^*$Authors contributed equally.}%
}

\begin{document}

\maketitle
\thispagestyle{empty}
\pagestyle{empty}


\begin{abstract}
Embodied agents need to plan and act reliably in real and complex 3D environments.
Classical planning (e.g., PDDL) offers structure and guarantees, but in practice it fails under noisy perception and incorrect predicate grounding. 
On the other hand, Large Language Models (LLMs)-based planners leverage commonsense reasoning, yet frequently propose actions that are unfeasible or unsafe. 
Following recent works that combine the two approaches, we introduce ContextMatters, a framework that fuses LLMs and classical planning to perform hierarchical goal relaxation: the LLM helps ground symbols to the scene and, when the target is unreachable, it proposes functionally equivalent goals that progressively relax constraints, adapting the goal to the context of the agent's environment. 
Operating on 3D Scene Graphs, this mechanism turns many nominally unfeasible tasks into tractable plans and enables context-aware partial achievement when full completion is not achievable.
Our experimental results show a +52.45\% Success Rate improvement over state-of-the-art LLMs+PDDL baselines, demonstrating the effectiveness of our approach. Moreover, we validate the execution of ContextMatters in a real-world scenario by deploying it on a TIAGo robot.
Code, dataset, and supplementary materials are available to the community at https://lab-rococo-sapienza.github.io/context-matters/.

\end{abstract}

\section{Introduction}

\begin{figure}[t!]
\centering\includegraphics[width=\linewidth]{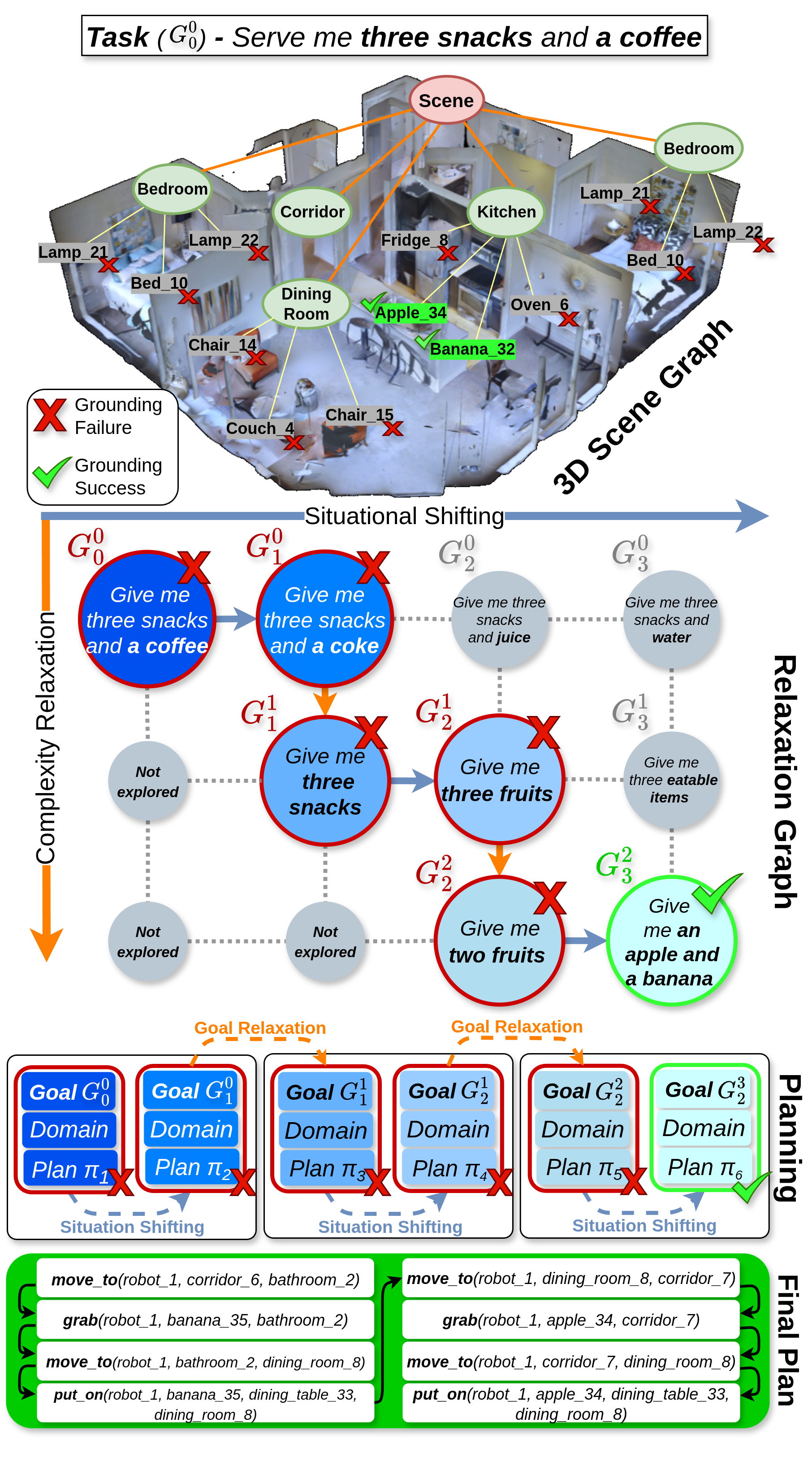}    \caption{Our architecture takes as input a 3D Scene Graph \cite{armeni20193d} of the environment and a task expressed in natural language. Unfeasible goals can be relaxed up to a certain degree into semantically similar ones, computing the corresponding plan. This mimics the capability of humans to change expected outcomes on-the-go depending on the \emph{context}.  \vspace{-0.5cm}}
\label{fig:intro}
\end{figure}

Planning with robots and embodied agents requires generating action sequences that an agent must execute to achieve a goal within its environment. This process becomes more challenging when moving from simulated to real-world settings, where robots can fail due to a failure in correctly capturing the world model in the plan’s preconditions. The two most common complementary task planning approaches can both break. Pure LLM planners~\cite{brown2020language,achiam2023gpt,touvron2023llama} leverage their commonsense knowledge to interpret intent \cite{li2024generative}, but often hallucinate missing preconditions and actions, yielding optimistic sequences that collapse at execution. On the other hand, pure PDDL planners~\cite{aeronautiques1998pddl} provide guarantees, yet treat unmet preconditions as dead-ends: if the goal, as represented in the planning domain, is unsatisfiable, planning fails, offering no principled way to adapt the goal while preserving intent.\looseness-1

Taking for instance a task ``\emph{set the dining table with two forks}'', carried out within a 3D environment, typically represented as a 3D Scene Graph (3DSG)~\cite{armeni20193d, Wald_2020_CVPR}, containing a dining table, cutlery drawers, a shelf, and a running dishwasher. In the actual scene, the top drawer is blocked and no clean forks are available, while spoons are reachable on the shelf. An LLM-only pipeline would often propose the idealized sequence ``open drawer $\rightarrow$ pick forks $\rightarrow$ place'', unless the blockage and unavailability are explicitly grounded; a PDDL-only pipeline correctly encodes those constraints and returns \emph{no plan}. Both these scenarios lack a mechanism treating failure as a cue to modify both \emph{what} to achieve and \emph{where/how} to achieve it, by adapting the user’s intent to the \emph{context} of the scene without discarding it.\looseness-1

We therefore ask the following research question: instead of failing, can an agent intelligently analyze its 3D environment to relax the goal into {\emph{functionally equivalent but contextually achievable objectives}?\looseness-1

To this end, we introduce \cm, a bidimensional relaxation architecture that jointly searches over \emph{functionality} (\emph{what to} achieve, at varying levels of semantic equivalence) and \emph{feasibility} (\emph{where/how} to achieve it under symbolic and physical constraints) to convert failure signals into intent-preserving, executable goals. Consider the example depicted in Fig.~\ref{fig:intro}, where the user requests ``\emph{serve me three snacks and a coffee}'' in a home represented as a 3D Scene Graph. The scene makes the original specification difficult since some beverages are unavailable while only a subset of snacks (\eg \texttt{banana\_32}, \texttt{apple\_34}) are reachable. \cm\ explores nearby goal variants such as ``\emph{three snacks and water/juice}'', ``\emph{two fruits}'', or ``\emph{three edible items}'', and validates each candidate with a classical planner until a concrete plan emerges. By employing a bidimensional relaxation, \cm\ proposes the \emph{minimal} scene-grounded modification that preserves intent while ensuring preconditions hold, converging on an executable plan when a relaxed goal satisfies feasibility checks. By comparing our approach with state-of-the-art LLM+PDDL planners on 3DSGs, we show the effectiveness of our architecture achieving +52.45\% improvement, while also executing it on a real robot setup. \looseness-1
To support research in this direction, we also provide a dataset of tasks that require some degree of relaxation in order to be correctly carried out in the environment.

To summarize, our main contributions are:
\begin{itemize}
  \item a novel contextual goal-relaxation formalism that reasons along two axes (functionality and feasibility) to preserve user intent while yielding executable goals.
  \item \cm, a planning framework that couples LLM commonsense for goal proposal with classical planning for feasibility validation and plan synthesis.
  \item a new dataset of 141 relaxation-prone tasks compatible with popular 3D environments and 3DSGs~\cite{armeni20193d,xiazamirhe2018gibsonenv}.
  \item an empirical evaluation on 3DSG planning benchmarks and a real-world demonstration on a TIAGo robot.
\end{itemize}

The remainder of this paper is structured as follows. Section~\ref{sec:relwork} reviews related work on robot planning using LLMs and classical methods, while Section~ \ref{sec:methodology} details the proposed approach. Experimental results are presented in Section~\ref{sec:results}, and conclusions are drawn in Section~\ref{sec:conclusions}.

\section{Related Work}
\label{sec:relwork}
\textbf{3D Scene Graphs.}
3D Scene Graphs~\cite{armeni20193d, Wald_2020_CVPR} have recently emerged as a versatile representation for indoor and outdoor environments~\cite{hughes2022hydrarealtimespatialperception, strader2024indooroutdoor3dscene}. Graph nodes typically refer to scene objects and their attributes (like materials, or affordances); graph edges denote spatial and semantic relations between the primitives of the environment. Their compact structure allows prompting a whole scene to an LLM for querying, facilitating a variety of downstream applications, like navigation in the environment~\cite{Werby_2024}, manipulation~\cite{gu2023conceptgraphs}, and task planning~\cite{rana2023sayplan, agia2022taskographyevaluatingrobottask}.\looseness-1

\textbf{LLMs as Planners.}
LLMs are increasingly used for planning in embodied agents. In the seminal SayCan framework~\cite{ahn2022can}, a robot couples its observations and affordances with an LLM to ground high-level tasks in real settings, exploiting the model’s semantic priors. Subsequent work uses LLMs as planners: with careful prompting, they function as zero-shot planners~\cite{huang2022language, song2023llmplanner}; with lightweight few-shot fine-tuning, they outperform state-of-the-art Vision-Language Navigation (VLN) models trained on larger datasets, benefiting from intrinsic commonsense~\cite{song2023llmplanner,dorbala2023can}. Performance improves further when inputs adopt structured representations, such as 3DSGs, which inject additional semantic context to the scene~\cite{rana2023sayplan}. Advances in open-vocabulary perception have broadened real-world applicability by enabling flexible grounding of novel objects and scenes~\cite{10802251, yang2023llmgrounderopenvocabulary3dvisual}. Finally, multi-level frameworks integrate reasoning, planning, and motion generation within a single LLM-centric loop~\cite{joublin2024copal}. Despite these improvements, robust long-horizon planning still remains a key open challenge.\looseness-1

\textbf{LLMs and Classical Planning.}
Combining LLMs with classical planners enables solving long-horizon tasks more reliably than pure LLM planning. For example, Ding et~al.~\cite{ding2023integrating} dynamically augment the robot’s action knowledge and modify the preconditions and effects of actions by leveraging task-oriented commonsense knowledge.
However, these approaches struggle in completely unseen environments, as they rely on a strict definition of both domain and problem, often leading to failures in these cases. To combine the strengths of natural language reasoning and classical planning, Liu~et~al.~\cite{liu2023llm+} propose a method that translates a natural language description of a planning problem into a PDDL representation, allowing it to be solved using a PDDL planner. Since LLMs can still hallucinate and generate infeasible plans due to incomplete domain knowledge, Liu et~al.~\cite{liu2024delta} introduce DELTA, an LLM-guided task planning framework. DELTA leverages 3D Scene Graphs as environmental representations within LLMs to quickly generate planning problem descriptions. To further improve planning efficiency, it breaks down long-term task goals into an autoregressive sequence of sub-goals using LLMs, enabling automated planners to solve complex tasks more effectively.      

DELTA is the current state-of-the-art in combining LLMs with structured scene representations for task planning. Its main limitation is its exclusive reliance on LLMs to produce a directly usable symbolic domain. Generation inaccuracies in this phase, such as invalid terms, predicates or grounding mismatches, can lead to plans that are not executable in the modeled environment. \cm \xspace addresses these issues through iterative domain validation, using a symbolic PDDL validator to provide feedback for its refinement.

\looseness-1

\section{Methodology}
\label{sec:methodology}
This section is organized as follows. Section \ref{subsec:taskadaptation} provides a formal definition of our planning approach, and introduces the \emph{shifting} and \emph{relaxation} operators. Section \ref{subsec:architecture} highlights the architecture of our context-augmented planning system. Lastly, Section \ref{subsec:implementation} describes the implementation of the system.

\subsection{Contextual Task Adaptation}
\label{subsec:taskadaptation}
\subsubsection*{\textbf{Stateful domain specification}} We use a PDDL-like \emph{stateful domain} $\Sigma \;=\; \langle \textit{Obj},\ \textit{Pred},\ \textit{Act},\ \textit{Init}\rangle$, where:
\begin{description}
  \item[\textit{Obj:}] finite set of \emph{typed object constants} (\eg \ttb{cup\_1:cup}, \ttb{table\_3:table}, \ttb{kitchen:room});
  \item[\textit{Pred}:] set of \emph{predicate symbols} with arities and type signatures; may be partitioned into \emph{static} (never change) and \emph{fluent} (state-dependent) predicates, and may include \emph{derived predicates};
  \item[\textit{Act}:] set of \emph{action schemas} (operator templates) with parameters, preconditions, and effects;
  \item[\textit{Init}:] the \emph{initial state} at \(t{=}0\), a set of true ground atoms over \textit{Obj} and \textit{Pred}.
\end{description}

\subsubsection*{\textbf{Representation mapping}}
Given as input a 3DSG of a scene $\ssem$, the function \(\mathcal{M}_{\text{repr}}\) maps $\ssem$, a knowledge base \(\mathcal{K}\), and an action library \(\mathcal{L}_{\text{act}}\) to a domain specification:
\[
\mathcal{M}_{\text{repr}}(\ssem,\mathcal{K},\mathcal{L}_{\text{act}})\ \mapsto\ 
\Sigma=\langle \textit{Obj},\textit{Pred},\textit{Act},\textit{Init}\rangle .
\]
In practice: \textit{Obj} is derived from typed instances in $\ssem$; \textit{Pred} from relations/attributes plus \(\mathcal{K}\); \textit{Act} from \(\mathcal{L}_{\text{act}}\); \textit{Init} from facts extracted/derived from \(\ssem\) and \(\mathcal{K}\). From now on, \(\mathcal{M}_{\text{repr}}(\ssem)\) will be shorthand for
\(\mathcal{M}_{\text{repr}}(\ssem, \mathcal{K}, \mathcal{L}_{\text{act}})\) .
\subsubsection*{\textbf{From scene to planning problem}}
Given a semantic environment \(\ssem\) and an initial goal \(G_0\), the planning problem is then defined as the tuple 
\[\mathcal{P}_0 \;=\; \langle \Sigma_0,\ G_0\rangle, \]
where 
\(
\Sigma_0 \;=\; \mathcal{M}_{\text{repr}}\!\big(\ssem\big)
\;=\; \langle \textit{Obj}_0,\ \textit{Pred}_0,\ \textit{Act},\ \textit{Init}_0\rangle
\) represents the initial planning domain specification induced by the environment.
\subsubsection*{\textbf{\textit{Situational shifts}}}
The \emph{situational shift operator}
\[
\Sigma_k \;=\; \Gamma_{\text{shift}}\!\big(\Sigma_{k-1},\, G_0,\, \ssem,\, \mathcal{M}_{\text{repr}}\big),
\quad k \ge 1,
\]
adapts the agent's understanding of the operating environment to the goal \(G_0\).
Assuming the set of actions \(\textit{Act}\) does not change, each shifted specification has the form
\[
\Sigma_k \;=\; \langle \textit{Obj}_k,\ \textit{Pred}_k,\ \textit{Act},\ \textit{Init}_k\rangle.
\]

The components \(Obj_k\) and \(Init_k\) are adapted to be relevant for the shifted planning problem after the shifting. We denote the \(k\)-fold application of the situational shift by \(\Gamma_{\text{shift}}^{\,k}\!\big(\Sigma_0,\, G_0,\, \ssem,\, \mathcal{M}_{\text{repr}}\big)\).
Similarly, a \emph{goal-shift operator}
\[
G_k \;=\; \Gamma_{\text{goal}}\!\big(G_{k-1},\, \ssem,\, \mathcal{M}_{\text{repr}}\big),
\quad k \ge 1,
\]
produces reformulations of the original intent over the current domain vocabulary.
We write \(G_k \sim G_0\) when \(G_k\) preserves the intent of \(G_0\) under the vocabulary of \(\Sigma_k\), \ie \( \Sigma_k \models G_k \) and \(G_k\) is logically interchangeable with \(G_0\) up to object renaming and available predicates induced by \(\mathcal{M}_{\text{repr}}\).

\subsubsection*{\textbf{Relaxation operator}}
Fix a domain specification \(\Sigma\) (\eg \(\Sigma_0\) or the current \(\Sigma_k\)).
For goal formulas interpreted over \(\Sigma\), let
\(\operatorname{Mod}_\Sigma(G) = \{\,s \mid s \models G\,\}\) denote the set of states that satisfy \(G\). We define the \emph{relaxation preorder} \(\succeq_{\text{rel}}\) by
\[
G' \succeq_{\text{rel}} G
\quad\Longleftrightarrow\quad
\operatorname{Mod}_\Sigma(G) \subseteq \operatorname{Mod}_\Sigma(G') .
\]
We write \(G' \succ_{\text{rel}} G\) when the inclusion is strict. Intuitively, \(G'\) is
a \emph{weaker / more general / more abstract} goal than \(G\).
A \emph{relaxation operator}
\[
G^{i} \;=\; \Delta_{\text{rel}}\!\big(G^{i-1}, \Sigma\big),
\quad i \ge 1,
\]
is \emph{valid} if it is monotone w.r.t.\ \(\succeq_{\text{rel}}\), \ie
\(G^{i} \succeq_{\text{rel}} G^{i-1}\).
By iterating, it induces a hierarchy
\[
G^{m} \succeq_{\text{rel}} \cdots \succeq_{\text{rel}}G^{1} \succeq_{\text{rel}} G^{0},
\]
or, in strict form when each step truly relaxes the goal,
\[
G^{m} \succ_{\text{rel}} \cdots \succ_{\text{rel}} G^{1} \succ_{\text{rel}} G^{0}.
\]
We denote the \(i\)-fold application of the relaxation operator by \(\Delta_{\text{rel}}^{\,i}(G^{0}, \Sigma)\).\looseness-1

According to these definitions, some relaxation examples are: 
(i) drop conjuncts (from \(\land\)-goals); (ii) replace constants with types or sets
(\eg \(\texttt{cup\_3}\) \(\to\) \(\exists x{:}\texttt{cup}\)); (iii) generalize predicates via \(\mathcal{K}\)
(\eg \(\texttt{on}\) \(\to\) \(\texttt{supported\_by}\)); (iv) widen numeric thresholds; (v) introduce disjunctions.
Each increases \(\operatorname{Mod}_\Sigma(\cdot)\) and thus respects \(\succeq_{\text{rel}}\).

\begin{figure*}[t!]
    \centering
    \includegraphics[width=1.0\linewidth]{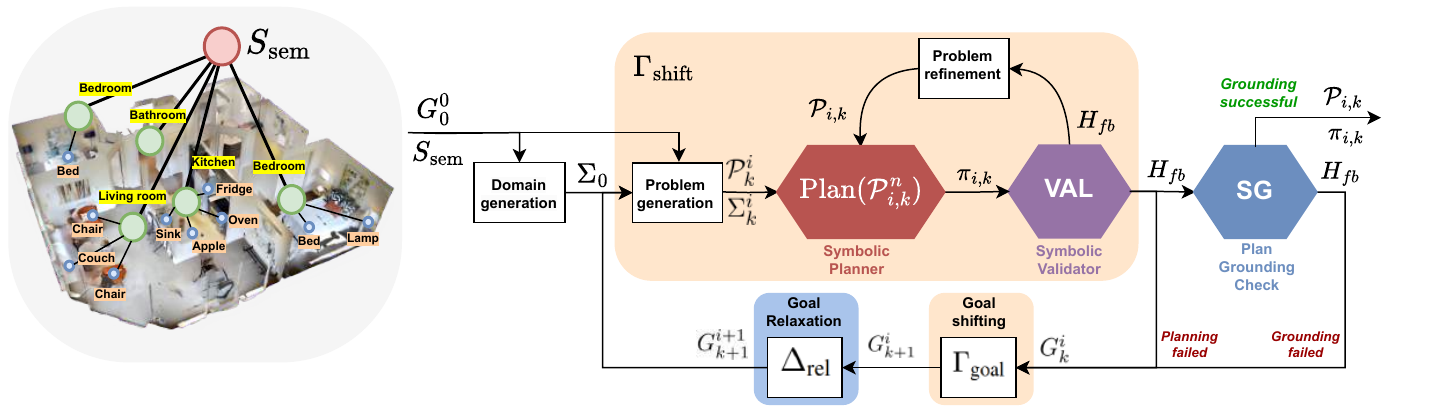}
    \caption{Proposed architecture. The generated domain is followed by a planning attempt. In case of failure, domain refinement follows the feedback of a symbolic validator (purple hexagon) and a grounding verifier (blue hexagon), mapping domain elements to real objects in the 3DSG. If the refinement loop is unsuccessful, an attempt is made to \emph{shift} the goal (orange block) and to optionally \emph{relax} it (blue block) to a more feasible but still intent-preserving version.}
    \vspace{-0.5cm}
    \label{fig:architecture}
\end{figure*}

\subsubsection*{\textbf{Combining the two operators}}
The application of a \textit{shift} \(\Gamma_{\text{shift}}^{\,k}\) followed by a \textit{relaxation} \(\Delta_{\text{rel}}^{\,i}\) to the initial pair \(\langle \Sigma_0, G_0^0 \rangle\) yields a shifted planning domain \(\Sigma_{k}\) and a shifted-relaxed goal \(G^{i}_{k}\), obtained by adapting the agent's understanding of the operating environment and then relaxing the goal within the shifted domain. \looseness-1 Formally, we first compute the \(k\)-fold shift
\[
\Sigma_{k} \;=\; \Gamma_{\text{shift}}^{\,k}\!\big(\Sigma_{0},\, G_{0}^0,\, \ssem,\, \mathcal{M}_{\text{repr}}\big),
\]
(optionally aligning the goal to the new vocabulary as \(G_{k}^0=\Gamma_{\text{goal}}^{\,k}(G_{0}^0,\ssem,\mathcal{M}_{\text{repr}})\), otherwise \(G_{k}^0\equiv G_{0}^0\)),
and then build the \(i\)-fold relaxation \emph{within} \(\Sigma_{k}\):
\[
G_k^i \;=\; \Delta_{\text{rel}}^{\,i}\!\big(G^{0}_{k},\, \Sigma_{k}\big), \qquad i \ge 0.
\]

We can now form a two-dimensional family of planning problems by combining each shifted domain \(\Sigma_{k}\) and its relaxed goals \(G^{i}_{k}\), whose solution is a plan \(\pi_{i,k}\):
\[
\mathcal{P}_{i,k} \;=\; \langle \Sigma_{k},\, G^{i}_{k} \rangle,
\qquad \forall\, k \in \{0,\dots,N\},\ \forall\, i \in \{0,\dots,M\}.
\]
Hence, each \(\mathcal{P}_{i,k}\) is a distinct planning problem with (i) domain specification \(\Sigma_{k}\) and (ii) goal specification \(G^{i}_{k}\); a plan \(\pi_{i,k}\) must satisfy \(G^{i}_{k}\) \emph{within} \(\Sigma_{k}\).\looseness-1

\subsubsection*{\textbf{Relaxation graph}}
For each pair \((i,k)\) we consider the shifted domain \(\Sigma_{k}\) and the shifted–relaxed goal \(G^{i}_{k}\), yielding
\[
\mathcal{P}_{i,k} \;=\; \langle \Sigma_{k},\, G^{i}_{k} \rangle,
\qquad
\pi_{i,k} \;=\; \operatorname{Plan}\!\bigl(\Sigma_{k},\, G^{i}_{k}\bigr).
\]
We define the \emph{relaxation graph} as the directed graph
\[
\mathcal{G}_{\text{relax}} \;=\; (V,E),\quad
V=\{\mathcal{P}_{i,k} \mid i=0{:}M,\; k=0{:}N\},
\]
with horizontal (shift) edges
\[
E_{\text{shift}}=\{\, \mathcal{P}_{i,k} \to \mathcal{P}_{i,k+1} \mid i,k \,\},
\]
and vertical (relax) edges
\[
E_{\text{rel}}=\{\, \mathcal{P}_{i,k} \to \mathcal{P}_{i+1,k} \mid i,k \,\},
\]
such that \(E=E_{\text{shift}}\cup E_{\text{rel}}\), and root \(\mathcal{P}_{0,0}=\langle \Sigma_{0}, G^{0}_{0}\rangle\).
Any path in \(\mathcal{G}_{\text{relax}}\) encodes a sequence of shifts and relaxations explored in search of a solvable instance. Visually:
\[
\begin{aligned}
&\langle \Sigma_{0}, G^{0}_{0}\rangle \;\to\; \langle \Sigma_{1}, G^{0}_{1}\rangle \;\to\; \cdots \;\to\; \langle \Sigma_{N}, G^{0}_{N}\rangle,\\
&\langle \Sigma_{0}, G^{1}_{0}\rangle \;\to\; \langle \Sigma_{1}, G^{1}_{1}\rangle \;\to\; \langle \Sigma_{2}, G^{1}_{2}\rangle \;\to\; \cdots, \\
&\quad \vdots \\
&\langle \Sigma_{0}, G^{M}_{0}\rangle \;\to\; \langle \Sigma_{1}, G^{M}_{1}\rangle \;\to\; \cdots \;\to\; \langle \Sigma_{N}, G^{M}_{N}\rangle.
\end{aligned}
\]
Horizontally (increasing \(k\)), \(\Gamma_{\text{shift}}\) adapts the domain to the scene while preserving the original intent; vertically (increasing \(i\)), \(\Delta_{\text{rel}}\) weakens the goal within the fixed \(\Sigma_{k}\). We traverse this space by prioritizing first moving right (shifting) as long as planning shows progress (\eg more preconditions become satisfiable or search effort drops); if progress stalls, we step down (relax) to simplify the goal, repeating until some \(\mathcal{P}_{i,k}\) is solvable, returning \(\pi_{i,k}\). Because \(\Sigma_{k}\) is derived from the 3DSG, the plan is grounded to the robot's percepts.

\subsection{Architecture}
\label{subsec:architecture}
In order to implement the \emph{shift operators} \(\Gamma_{\text{shift}}\) and \(\Gamma_{\text{goal}}\), producing respectively a new shifted domain and goal, and the \textit{relaxation operator} \( \Delta_{\text{rel}} \), producing a relaxed goal, the architecture (see Fig. \ref{fig:architecture}) exploits commonsense reasoning of LLMs, through the intrinsic representation $\mathcal{M}_{\text{repr}}$. The agent's operating environment is initially modeled as a 3DSG \( \ssem \), then filtered by keeping only information about the location of objects in the scene, each tagged with a brief description. \looseness-1
\subsubsection*{\textbf{Domain generation}}
A symbolic domain \( \Sigma_0 \), based on the agent's skills and the environment, can be either provided as a structured prior, or generated from \( \ssem \), the initial goal \( G_0 \), a knowledge base $\mathcal{K}$ and the action library $\mathcal{L}_{\text{act}}$. \looseness-1
%
%
%

\subsubsection*{\textbf{Iterative problem refinement}}
The $\Gamma_{\text{shift}}$ operator generates a shifted planning problem $\mathcal{P}_{i,k}$ from the domain $\Sigma_0$, the current goal $G^i_k$ and the 3DSG $S_{\text{sem}}$.
$\mathcal{P}_{i,k}$ might not be immediately solvable, due to hallucinations introduced in the domain generation or misalignments between any element of the tuple $\langle Obj, Pred, Act, Init \rangle$ and the ground information contained in $\ssem$. This would make it impossible to produce an executable plan, namely a plan $\pi_{i,k}$ that is groundable in the scene graph $\ssem$. Therefore, through \textit{Problem refinement}, the generated problem is corrected by maximizing its semantic relevance to the current goal and its alignment to $\ssem$. The refinement is iterated in case a solution plan is not found. The set $H_{fb}$, containing natural language feedback obtained throughout the symbolic planning process, is used to guide the refinement, until a plan can be computed.
We can consider the \textit{situational shifting operator} $\Gamma_{\text{shift}}$, as the composition of problem generation and iterative refinement. \looseness-1
At this stage, the goal is not shifted yet. The planning process will conservatively attempt to modify the representation of the operating environment without modifying the goal, if possible, to avoid any unnecessary variation with respect to the original task.  

\subsubsection*{\textbf{Plan grounding check}} If a successful plan \( \pi_{i,k} \) is obtained, its correctness is evaluated by verifying the grounding of each action parameter, in an attempt to find a mapping between elements in domain \( \Sigma^i_k \) and the environment representation \( S_{\text{sem}} \). If this process is successful, the plan is accepted. \looseness-1
\subsubsection*{\textbf{Goal shifting and relaxation}}
If even after the iterative refinement it is still impossible to compute a groundable plan \( \pi_{i,k} \) for the current planning problem \( \mathcal{P}_{i,k} \), the \textit{situational shifting} proceeds with a \textit{goal shifting} step, followed by a \textit{goal relaxation} \( G^{i+1}_{k+1} = (\Gamma^k_{\text{goal}} \circ \Delta_{\text{rel}})(G^i_k)\), producing a relaxed planning problem \( \mathcal{P}_{i+1,k+1} \).
The generation, refinement, grounding and relaxation process is then iteratively repeated, until a groundable plan is computed, or a maximum number of attempts has been reached. As shown in Fig.~\ref{fig:architecture}, the \textit{shifting operator} $\Gamma_{\text{shift}}$ is the composition of the domain generation, problem shifting sub-step and goal shifting sub-steps.

\begin{algorithm}[t]
\small
\caption{\textit{Implementation of the architecture}.}
\label{alg:workflow}

\KwIn{$G^0_0$; Semantically enhanced 3DSG $\ssem$;  Knowledge base $\mathcal{K}$; Action Library $\mathcal{L}_{\text{act}}$}
\KwOut{Executable plan $\pi_{i,k}$; Final problem $\mathcal{P}_{i,k}$}

{$\Sigma_0 \gets LLM\!\left(G_0,\ssem,\Sigma_{\text{desc}}\right)$} \hfill\textcolor{gray}{\Comment{Domain generation}}
\\
$i,k \gets 0$; $groundOK \gets False$; \\
\While{$\pi_{i,k}=\emptyset \land \neg groundOK$}{
\vbox{\colorbox{gammashiftcolor}{\vbox{
\textbf{\Comment{$\Gamma_{\text{shift}}$}}\\
  $\mathcal{P}_{i,k} \gets LLM\!\left(\Sigma^i_k,G^i_k,\ssem\right)$; \ \hfill\textcolor{gray}{\Comment{Probl. gen.}}\\
  \colorbox{plancolor}{\textcolor{white}{$\pi_{i,k} \gets \operatorname{Plan}\!\left(\mathcal{P}_{i,k}\right)$}}; \ \hfill\textcolor{gray}{\Comment{Symbolic Planner}}\\
  \While{$\pi_{i,k} =\emptyset$}{
    \colorbox{VALcolor}{\textcolor{white}{$H_{fb} \gets \operatorname{VAL}\!\left(\mathcal{P}_{i,k}\right)$}}; \ \hfill\textcolor{gray}{\Comment{Symbolic Validator}}\\ 
    $\mathcal{P}_{i,k} \gets LLM\!\left(G^i_k,\Sigma^i_k,\ssem,H_{fb}\right)$; \ \hfill\textcolor{gray}{\Comment{Probl. ref.}}\\
    \colorbox{plancolor}{\textcolor{white}{$\pi_{i,k} \gets \operatorname{Plan}\!\left(\mathcal{P}_{i,k}\right)$}}; \\
  }
}}}

    \If{$\pi_{i,k} \neq \emptyset$}{
        \colorbox{groundcolor}{\textcolor{white}{groundOK, $H_{fb} \gets \operatorname{SG}\!\left(\pi_{i,k},\ssem\right)$}}; \ \hfill\textcolor{gray}{\Comment{Grounding}} \\
        \If{groundOK}{
            \textbf{break};\\
        }
    }

  \colorbox{gammashiftcolor}{$G^i_{k+1} \gets LLM\!\left(G^i_k,\ssem\right)$}; \ \hfill\textbf{\Comment{$\Gamma_{\text{goal}}$}}\\
  \colorbox{relaxcolor}{$G^{i+1}_{k+1} \gets LLM\!\left(G^i_{k+1},\ssem \right)$}; \ \hfill\textbf{\Comment{$\Delta_{\text{rel}}$}}\\
  $i \gets i{+}1$; \ $k \gets k{+}1$\;
}
\KwRet{$\mathcal{P}_{i,k},\ \pi_{i,k}$}\;
\end{algorithm}

\subsection{Implementation}
\label{subsec:implementation}
Algorithm~\ref{alg:workflow} describes the high-level steps documented in Section~\ref{subsec:architecture}. In practice, they are realized as a set of LLM prompts, translating the abstract architecture into concrete instructions, containing \boldred{constraints}, \boldblue{data} or \boldblack{examples}. Full prompts are available in the provided repository.

The raw 3DSG undergoes \textit{Context Distillation} in all LLM prompts, where it is pre-processed into a distilled semantic representation \(S_{\text{sem}}\), optimizing the LLM context window usage, by retaining only relevant data. Algorithm~\ref{alg:workflow} comprises two nested loops: the outer loop generates relaxed goals and candidate plans; the inner loop refines and validates them, or, if unsuccessful, triggers the next relaxation/shift.
\subsubsection*{\textbf{Domain generation}}
Initially, the \textbf{planning domain} $\Sigma_0$ is generated, by instructing the LLM to create a PDDL domain satisfying the initial goal $G^0_0$ in $\ssem$, with the prompt:
    \begin{tcolorbox}[
      colframe=black!75!black,
      boxsep=0pt,
      left=2pt,
      right=2pt,
      top=2pt,
      bottom=2pt,
      sharp corners,
      enhanced,
      breakable,
      fontupper=\small
    ]
      \colorbox{red!5}{\parbox{\dimexpr\linewidth-4pt\relax}{\textcolor{red!70!black}{\textbf{System Prompt:}} Given a description of the planning domain, the domain actions and the domain objects, you must generate a PDDL domain file. \boldred{[Generation constraints]},
%
%
    \boldred{[PDDL 1.2 Specifications]}, \boldblue{[Domain description]}, \boldblack{[PDDL domain example]}}}\\[-1pt]
      \colorbox{blue!5}{\parbox{\dimexpr\linewidth-4pt\relax}{\textcolor{blue!70!black}{\textbf{User Prompt:}} Extract object types and actions from the following description and generate a corresponding PDDL domain. \boldblue{[Goal]}. \boldblue{[Domain description]}. The generated PDDL domain incorporates these elements and respects the provided preconditions and effects. \boldred{[Generation constraints]}}}
    \end{tcolorbox}
    At every outer loop iteration, the LLM generates a PDDL problem matching the generated PDDL domain, as follows:
\begin{tcolorbox}[
      colframe=black!75!black,
      boxsep=0pt,
      left=2pt,
      right=2pt,
      top=2pt,
      bottom=2pt,
      sharp corners,
      enhanced,
      breakable,
      fontupper=\small
    ]
      \colorbox{red!5}{\parbox{\dimexpr\linewidth-4pt\relax}{\textcolor{red!70!black}{\textbf{System Prompt:}} Generate a PDDL problem file given: \boldred{[Description of the provided elements]}. 
      The environment is represented as a scene graph, with the following features: \boldred{[3DSG structure description]}. \boldred{[PDDL syntax constraints]}.
      }}\\[-1pt]
      \colorbox{blue!5}{\parbox{\dimexpr\linewidth-4pt\relax}{\textcolor{blue!70!black}{\textbf{User Prompt:}} \boldblue{[Goal]}, \boldblue{[PDDL domain]}, \boldblue{[Distilled 3DSG]}, \boldblack{[3DSG example]}, \boldblack{[PDDL domain/problem example]}}}
    \end{tcolorbox}

    \subsubsection*{\textbf{Iterative problem refinement}} If a plan cannot be found, the solution is validated by the symbolic validator, producing a set of natural language feedback $H_{fb}$ about PDDL correctness. The LLM is then prompted to reason about the cause of the error in a chain-of-thought, to correct the PDDL:
\begin{tcolorbox}[
      colframe=black!75!black,
      boxsep=0pt,
      left=2pt,
      right=2pt,
      top=2pt,
      bottom=2pt,
      sharp corners,
      enhanced,
      breakable,
      fontupper=\small
    ]
      \colorbox{red!5}{\parbox{\dimexpr\linewidth-4pt\relax}{\textcolor{red!70!black}{\textbf{System Prompt:}} Given the PDDL domain and problem, planner output, and scene, your job is to figure out why planning failed.
      }}\\[-1pt]
      \colorbox{blue!5}{\parbox{\dimexpr\linewidth-4pt\relax}{\textcolor{blue!70!black}{\textbf{User Prompt:}} \boldblue{[PDDL domain]}, \boldblue{[PDDL problem]}, \boldblue{[Goal]}, \boldblue{[Distilled 3DSG]}, \boldblue{[Planning/Validation/Grounding feedback]}. Please provide a clear explanation of the possible reason(s) for the planning failure. At the end provide detailed suggestions to solve the issues you found. \boldred{[Generation constraints]}}}
    \end{tcolorbox}

    The LLM is then instructed to properly correct the PDDL problem by integrating the analysis produced in the reasoning step and the feedback from the validation $H_{fb}$. while adhering to a subset of the PDDL 1.2 formalism~\cite{aeronautiques1998pddl}.

    \begin{tcolorbox}[
      colframe=black!75!black,
      boxsep=0pt,
      left=2pt,
      right=2pt,
      top=2pt,
      bottom=2pt,
      sharp corners,
      enhanced,
      breakable,
      fontupper=\small
    ]
      \colorbox{red!5}{\parbox{\dimexpr\linewidth-4pt\relax}{\textcolor{red!70!black}{\textbf{System Prompt:}} Given the PDDL domain, the previous PDDL problem, and a failure analysis, rewrite or fix the PDDL problem to address the failure according to the given suggestions.}}\\[-1pt]
      \colorbox{blue!5}{\parbox{\dimexpr\linewidth-4pt\relax}{\textcolor{blue!70!black}{\textbf{User Prompt:}} \boldblue{[PDDL problem]}, \boldblue{[Previous LLM output]}. (1) Fix domain-problem inconsistencies. (2) Match PDDL objects/types/predicates with the domain and (3) Ensure the goal is achievable. \boldblack{[Output format]}. \boldred{[Generation constraints]}.}}\\[-1pt]
    \end{tcolorbox}
    The two-step strategy improves the chances of success. 
    \subsubsection*{\textbf{Grounding check}} The correctness of a plan is verified by virtually executing the plan in the 3DSG, to detect LLM-induced hallucinations, in case objects not available in the original 3DSG are misplaced in the generated PDDL domain. In practice, this consists in a scene consistency check over the candidate symbolic plan: plans introducing objects having no correspondence in the actual 3DSG are rejected. Feedback obtained from this additional check is then added to the set $H_{fb}$, to correct the problem in subsequent iterations. 
    
    \subsubsection*{\textbf{Goal shifting and relaxation}} If a plan is not found, the goal must be reformulated through shifting and relaxation. The goal shifting step $\Gamma_{\text{goal}}$, prompts the LLM to reformulate the goal using alternative objects in the scene.
    \begin{tcolorbox}[
      colframe=black!75!black,
      boxsep=0pt,
      left=2pt,
      right=2pt,
      top=2pt,
      bottom=2pt,
      sharp corners,
      enhanced,
      breakable,
      fontupper=\small
    ]
      \colorbox{red!5}{\parbox{\dimexpr\linewidth-4pt\relax}{\textcolor{red!70!black}{\textbf{System Prompt:}} Identify the objects necessary to perform a task in a scene graph. Given a high-level task, your goal will be to identify similar objects, objects that can both be used for the same functions. \boldblack{[Execution example]}}}\\[-1pt]
      \colorbox{blue!5}{\parbox{\dimexpr\linewidth-4pt\relax}{\textcolor{blue!70!black}{\textbf{User Prompt:}} \boldblue{[Distilled 3DSG]}, \boldblue{[Goal]}}}
    \end{tcolorbox}
    The goal relaxation step $\Delta_{\text{rel}}$, is implemented by instructing the LLM to reason about how to decompose the goal in single steps and to remove implicit restrictions, if needed, or retain the previous formulation if it is already feasible, with the prompt:
\begin{tcolorbox}[
      colframe=black!75!black,
      boxsep=0pt,
      left=2pt,
      right=2pt,
      top=2pt,
      bottom=2pt,
      sharp corners,
      enhanced,
      breakable,
      fontupper=\small
    ]
      \colorbox{red!5}{\parbox{\dimexpr\linewidth-4pt\relax}{\textcolor{red!70!black}{\textbf{System Prompt:}} Given a task and a description of the available objects and their locations, determine whether the given objective is achievable or propose a relaxed objective, semantically similar to the original and still feasible, removing the least important restrictions first. \boldred{[Generation constraints]} \boldblack{[Execution example]}}}\\[-1pt]
      \colorbox{blue!5}{\parbox{\dimexpr\linewidth-4pt\relax}{\textcolor{blue!70!black}{\textbf{User Prompt:}} \boldblue{[Previous LLM response]}, \boldblue{[Goal]}.}}
    \end{tcolorbox}
As both steps are generative in nature, they can be considered LLM-based ``proposal" functions. Consequently, an invocation of a shifting or relaxation does not necessarily imply a modification of the goal; it represents an attempt to relax the objective conditioned on the available context.

\section{Experimental Results}
\label{sec:results}

In this section we discuss our experimental setup comprising baselines, dataset, metrics and main results (Sections \ref{subsec:baselines}, \ref{subsec:dataset}, \ref{subsec:metrics} and \ref{subsec:exps}). Additionally, we also demonstrate our approach on a real robot setup (Section \ref{subsec:robot}).

\subsection{Baselines}
\label{subsec:baselines}
We compare against three popular baselines: LLMAsPlanner, SayPlan, and DELTA.

\textbf{LLMAsPlanner} is a simple prompting baseline that provides the 3DSG of the environment and the natural-language goal to an LLM, acting as a planner.\looseness-1

\textbf{SayPlan}~\cite{rana2023sayplan} performs task planning on 3DSGs in a purely autoregressive manner, without an explicit symbolic planner. Because no official code was released, we reimplemented the method following the prompts and settings described in the paper in the same way as Liu et al.~\cite{liu2024delta}.\looseness-1

\textbf{DELTA}~\cite{liu2024delta} is the state-of-the-art system for task planning on 3DSGs. It combines LLMs with PDDL planning to decompose the original task into feasible sub-goals for efficient problem solving. We use the officially released codebase, integrated into our framework.\looseness-1

\subsection{Dataset}
\label{subsec:dataset}
We adopt the 3DSG dataset of Armeni et al.~\cite{armeni20193d}, providing a collection of 3DSGs extracted from multi-room environments~\cite{xiazamirhe2018gibsonenv}. For comparability we follow the same task domains and scenarios as DELTA \cite{liu2024delta}: environments \emph{Allensville, Parole, Shelbiana, Kemblesville} and tasks \textit{Laundry (LA), PC Assembly (PC), Dining Table Setup (DS), House Cleaning (HC), Home Office Setup (OS)}\footnote{For task definitions, see~\cite{liu2024delta}.}. 
 We augment the original 3DSG with objects relevant to the task. Most tasks are solvable with the added objects. To stress adaptation, we introduce an additional set of problem instances that are intrinsically impossible, because some necessary objects are removed in the augmentation process, while semantically related substitutes or weaker variants are made available. This set comprises tasks that are not immediately satisfiable, due to unavailability of key objects, therefore requiring \emph{shifts} or \emph{relaxations} of the original goal.
To assess the scalability of this approach, we introduce a \textit{General (GN)} benchmark covering six additional environments (\textit{Klickitat, Lakeville, Leonardo}, \textit{Lindenwood, Markleeville, Marstons}), featuring generic pick-and-place tasks such as ``\textit{move all the apples from the kitchen to the fruit bowl in the dining room}". In total, our benchmark spans 141 tasks across 10 environments, compared to DELTA's 15 tasks in 4 environments. \looseness-1
The official DELTA implementation augments 3DSGs with post-hoc attributes and states absent from the original dataset. For a fair comparison without extra information, we evaluate strictly on the unmodified dataset.

\begingroup
\renewcommand{\arraystretch}{1.2}

\begin{table}[t] 
\centering
\caption{Comparison of baselines and our approach, w/ and w/o domain generation and grounding (gr) in the environment.}
\label{tab:results_table}
\resizebox{0.49\textwidth}{!}{
\begin{tabular}{c c c c c c c}

\multicolumn{3}{c}{} & \textbf{\shortstack{SR (\%)\\Grounding +\\Planning}} & \textbf{\shortstack{SR (\%)\\Planning\\only}} & \textbf{\shortstack{Avg.\\Planning\\time (s)}} & \textbf{\shortstack{Avg.\\Plan\\Length}} \\ 
\hline
\hline

\multirow{6}{*}{\rotatebox{90}{\textbf{\shortstack{Domain \\ generation}}}} 
    & \multirow{2}{*}{\shortstack{\textit{CM }\\\textit{w/o Relaxation (ours)}}}
        \rule{0mm}{4mm}
    & \textit{Gr} & 66.94 & 88.15 & 38.68 & 15.72\\ 

    &  & \textit{w/o Gr} & - & - & - &  -\\
    
    & \multirow{2}{*}{\shortstack{\textbf{CM }\\\textbf{w/ Relaxation (ours)}}}
        \rule{0mm}{4mm}
    & \textbf{\textit{Gr}} & \textbf{91.73} & 95.52 & 19.0 &  17.35\\ 
    &  & \textit{w/o Gr} & - & \textbf{99.02} & 24.24 &  17.61\\
    
    & \multirow{2}{*}{\shortstack{DELTA~\cite{liu2024delta}}}
        \rule{0mm}{4mm}
    & \textit{Gr} & 	39.28 & 65.76 & \textbf{0.03} &  23.52 \\  

    &  & \textit{w/o Gr} & - & 49.14 & \textbf{0.03} & 17.12 \\  
\hline
        \rule{0mm}{4mm}
\multirow{6}{*}{\rotatebox{90}{\textbf{\shortstack{w/o \\Domain \\Generation} }}} 
    & \textit{CM} (ours) 
    & \textit{Gr} & \textbf{91.54} & \textbf{97.22} & 9.94 &  17.81\\ 

    & {\shortstack{DELTA~\cite{liu2024delta}}}
        \rule{0mm}{4mm}
    & \textit{Gr} & 13.89 & 18.86 & \textbf{0.02} & 11.74\\  

&  \\
    & {\shortstack{SayPlan~\cite{rana2023sayplan}}} 
        \rule{0mm}{2mm}
    & \textit{w/o Gr} & - & 46.0 & 28.15 & 19.6\\  

    &  \\
    
    & {\shortstack{LLMAsPlanner}} 
        \rule{0mm}{2mm}
    & \textit{w/o Gr} & - & 71.2 & 8.84 & 21.6\\  
\bottomrule
\end{tabular}
}
\end{table}
\endgroup

\subsection{Metrics}
\label{subsec:metrics}
The experimental evaluation focuses mainly on the \emph{Success Rate} (SR) of the planning and grounding process, as the ratio between the successful tasks and the total number of tasks. A task is considered solved if both the planning and the grounding in the 3DSG are successful. Table \ref{tab:results_table} also shows the average SR of the planning step alone (considering only successful planning) and the \emph{Average Planning Time}. 

\subsection{Experiments}
\label{subsec:exps}
We used \textit{GPT-4o} for all experiments, with temperature set to 0 to minimize randomness in the output. We validate the symbolic syntax of both the outputs of DELTA and \emph{ContextMatters} using VAL~\cite{howey2004validating}, and then with a grounding check. To this aim, the plan is virtually executed using PDDLGym~\cite{silver2020pddlgym}, ensuring that no object label or location hallucinations are introduced by the LLM. The end-to-end nature of methods like SayPlan and LLMAsPlanner prevents the use of formal validation tools. As a result, they cannot offer strong guarantees of plan correctness and feasibility, a key limitation our symbolic framework is designed to overcome. SayPlan and LLMAsPlanner cannot benefit from the same grounding check due to the lack of a symbolic planning domain.\looseness-1

\begin{figure*}[t]
    \centering
    \includegraphics[width=.72\linewidth]{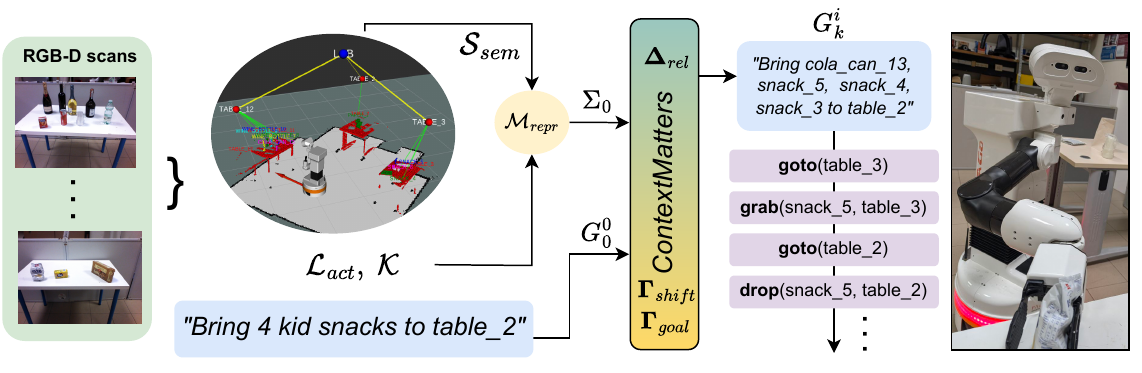}
    \caption{Pipeline employed to deploy \cm\ on a real robot. From perception we build a 3DSG of the environment, fed into our pipeline with the initial goal. We obtain the feasible goal in output, which is then translated into a sequence of groundable actions, executed by our robot.}
    \vspace{-0.4cm}
    \label{fig:realexperiment}
\end{figure*}
\begin{figure}[t]
    \centering
    \includegraphics[width=1.0\linewidth]{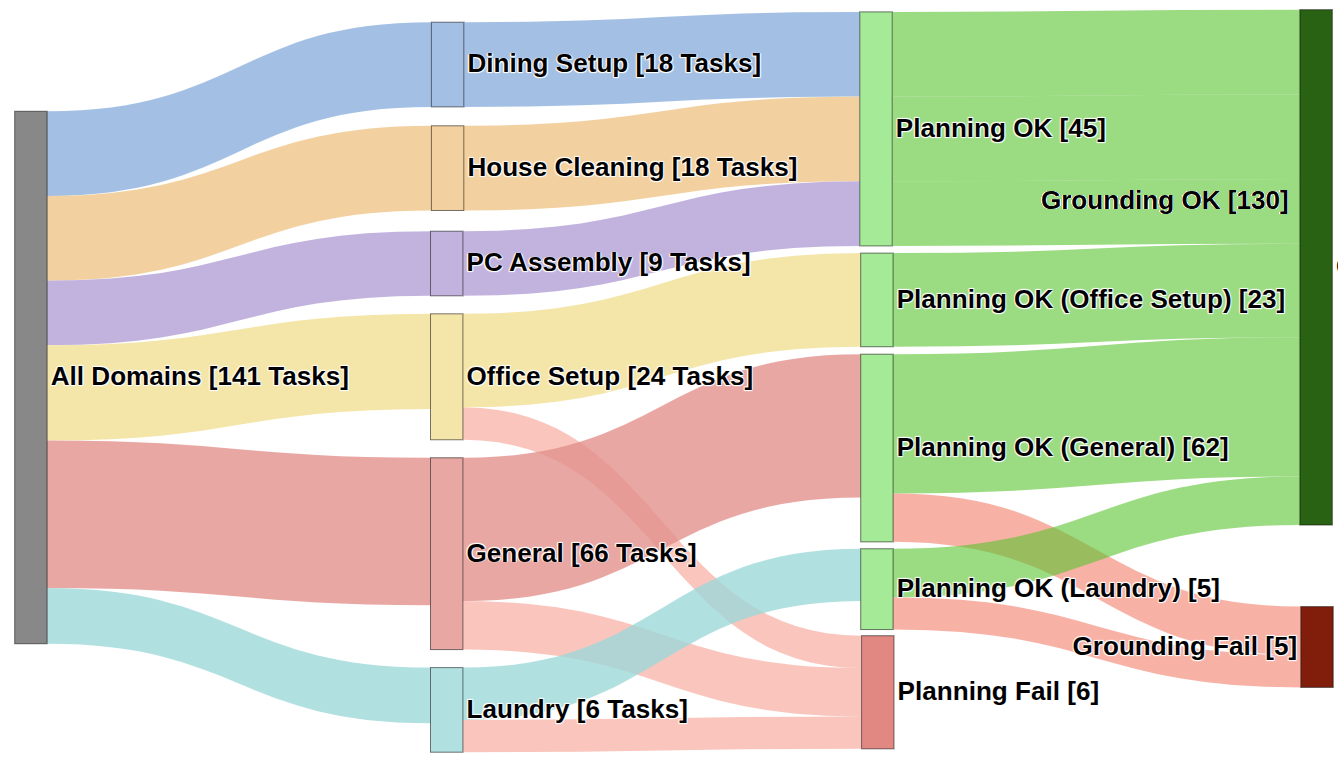}
    \caption{Results of \cm\ across our six-domain dataset, evaluated without domain generation and with grounding on the scene graph.}
    \vspace{-0.4cm}
    \label{fig:diagramCM}
\end{figure}

\subsection{Discussion} 
Table \ref{tab:results_table} shows that \cm\ achieves the best results on all tasks, both with and without domain generation. 
The LLMAsPlanner and SayPlan baselines show how LLMs are not directly applicable to the generation of groundable symbolic plans: in a more structured approach, the LLM is used to model the planning domain, then used to compute the plan. Our main architecture (in bold) features both domain generation and grounding checks, providing feedback to relax and refine the PDDL problem, for better feasibility of the computed plan.
Experiments show that relaxations, intended as any combination of goal shifting and semantic relaxation, improve the SR compared to open-loop architectures. The first row of Table \ref{tab:results_table} shows that both the Planning and Grounding SR of \cm\ decrease without goal relaxations. Following Algorithm \ref{alg:workflow}, more relaxations translate to more allowed iterations of the outer loop, progressively shifting the planning domain and adapting it to the available objects in the 3DSG, improving the chances of correcting any syntax or modeling issues in the PDDL domain formulation.
While DELTA computes the SR only on successfully generated plans, assuming that the 3DSG is faithfully reproduced throughout the PDDL generation, we add a grounding check, verifying that plan actions reference existing domain objects in the 3DSG. 
In case of failure, grounding errors provide corrective feedback to the following iterations. While DELTA optimizes planning time by decomposing the planning problem into sub-problems, its performance in the grounding step is notably worse due to the lack of a self-supervised PDDL refinement loop, which proved essential in \cm\ for correcting PDDL syntax errors and hallucinations.
%
%
%
%
%
%
 %
%
%
%
%
%
%
The Sankey diagram in Fig.~\ref{fig:diagramCM} shows the performance across all 141 tasks, highlighting overall end-to-end success rates of \cm\ while also clearly isolating failure modes.\looseness-1 
Table \ref{tab:planning_statistics} shows the performances of the main architecture on each task, ordered by difficulty, with several statistics, including time spent for LLM inference over a problem and the corresponding Planning SR. Easier tasks, requiring simple pick-move-place actions, consist in collecting objects scattered in the 3DSG, therefore the planner tends to produce longer plans by searching in a broader space, as shown by the longer planning time and the higher number of expanded nodes along the search tree. The two most difficult tasks, Office Setup and Laundry, require better domain modeling by the LLM, and better adaptation, causing longer inference times, while the required planning effort is lower.


\begingroup
\renewcommand{\arraystretch}{1.4}

\begin{table}[t] 
\centering

\caption{Average performance on the General (GN), House Cleaning (HC), PC Assembly (PC), Office Setup (OS), Laundry (LA), and Dining Setup (DS)  tasks.}
\label{tab:planning_statistics}
\resizebox{0.49\textwidth}{!}{
\begin{tabular}{c c c c c c c}

\multicolumn{1}{c}{} & \textbf{\shortstack{SR (\%)}} & \textbf{\shortstack{Plan \\Length}} & \textbf{\shortstack{Planning \\Time (s)}} & \textbf{\shortstack{Expanded\\ Nodes}} & \textbf{\shortstack{Inference \\Time (s)}}\\ 

\hline
\hline  

 \textbf{\shortstack{HC}}
     & 100.0 & 17.72 & 16.93 & 3971.83 & 57.86\\  

\textbf{\shortstack{PC}}
     & 100.0 & 23.22 & 9.5 & 832689.56 & 33.18 \\  

 \textbf{\shortstack{DS}}
     & 100.0 & 27.22 & 53.71 & 1237733.28 & 89.12 \\
     
\textbf{\shortstack{GN}}
     & 93.55 & 15.23 & 15.01 & 15441.76 & 54.9\\  

 \textbf{\shortstack{LA}}
     & 83.33 & 15.0 & 14.13 & 15039.0 & 169.78 \\ 
     
 \textbf{\shortstack{OS}}
     & 95.83 & 5.7 & 4.7 & 80.96  & 217.14 \\  

 \bottomrule
 \end{tabular}
}
 \end{table}
\endgroup

\subsection{Real robot setup}
\label{subsec:robot}
We tested the architecture on a real TIAGo robot, serving food within a real environment \footnote{https://pal-robotics.com/robot/tiago/}. The 3DSG was generated using the EMPOWER architecture~\cite{10802251}, which uses RGB-D images to reproject semantic information from the camera into the 3D environment. Panoptic masks on detected point clouds are used to create a 3DSG following the dataset structure of~\cite{armeni20193d} (Fig.~\ref{fig:realexperiment}).
We employed \cm\ without domain generation, with the scene graph grounding check, enabling planning in the real environment with plan feasibility constraints. The domain description was modeled to map the high-level PDDL actions to the physical capabilities of the TIAGo robot. (\eg manipulating objects with the gripper). The robot arm is controlled with the \emph{MoveIt!} motion planning framework~\cite{coleman2014reducing}.
The experiment was conducted with the natural language task: \textit{``Bring 4 kid snacks to \texttt{table\_2}"}, appropriately relaxed to: \textit{``Bring \texttt{cola\_can\_13}, \texttt{snack\_5}, \texttt{snack\_4}, \texttt{snack\_3} to \texttt{table\_2}"}. Given that only three snacks were available in the environment, goal shifting replaced one snack with a cola can (as it is common sense that wine bottles cannot be served to kids). The resulting plan is correctly executable in the real environment, successfully moving 
the required objects from \texttt{table\_3} to \texttt{table\_2}. Full execution is described in the supplementary material.

\section{Conclusion}
\label{sec:conclusions}

In this work, we presented \cm, a novel framework addressing a fundamental challenge in embodied AI planning: the gap between user intent and environmental constraints. Our approach introduces a bidimensional relaxation mechanism that systematically searches across both functional equivalence and feasibility dimensions, enabling robots to adapt goals to their operational context while preserving task semantics.\looseness-1

Our key contribution lies in the formalization of contextual goal relaxation through the integration of situational shift ($\Gamma_{\text{shift}}$) and relaxation ($\Delta_{\text{rel}}$) operators, enabling principled adaptation when exact goal satisfaction proves impossible. By combining the commonsense reasoning capabilities of Large Language Models with the formal guarantees of classical PDDL planning, \cm\ demonstrates that seemingly infeasible tasks can often be transformed into executable plans through intelligent goal adaptation.\looseness-1

Quantitative results show that our approach achieves a substantial improvement over state-of-the-art methods, and the successful deployment on a TIAGo robot further validates the practical applicability of our framework in real-world scenarios, where perfect environmental conditions rarely align with idealized task specifications.\looseness-1

We believe that the ability to intelligently relax and adapt goals represents a crucial step toward truly robust embodied AI systems. As we continue to deploy robots in unstructured, real-world environments, the capacity to reason about what can be achieved given the context, rather than failing on unmet preconditions, will prove essential for practical autonomy.\looseness-1

\section*{Acknowledgements}
This work has been carried out while Emanuele Musumeci, Michele Brienza and Francesco Argenziano were enrolled in the Italian National Doctorate on Artificial Intelligence run by Sapienza University of Rome. Michele Brienza is funded by the European Union - Next Generation EU, Mission I.4.1 Borse PNRR Pubblica Amministrazione (Missione 4) Component 1 CUP B53C23003540006. This work has been partially supported by PNRR MUR project PE0000013-FAIR.

\bibliographystyle{IEEEtran}
\bibliography{references_short.bib}

\end{document}